**The Way We Prompt: Conceptual Blending, Neural Dynamics, and Prompt-Induced Transitions in LLMs**

Makoto Sato

Mathematical Neuroscience Unit, Institute for Frontier Science Initiative, Laboratory of Developmental Neurobiology, Graduate School of Medical Sciences, Kanazawa University, Kanazawa, Ishikawa, Japan.

Correspondence: makotos@staff.kanazawa-u.ac.jp

**Abstract**

Large language models (LLMs), inspired by neuroscience, exhibit behaviors that often evoke a sense of personality and intelligence—yet the mechanisms behind these effects remain elusive. Here, we operationalize Conceptual Blending Theory (CBT) as an experimental framework, using prompt-based methods to reveal how LLMs blend and compress meaning. By systematically investigating Prompt-Induced Transitions (PIT) and Prompt-Induced Hallucinations (PIH), we uncover structural parallels and divergences between artificial and biological cognition. Our approach bridges linguistics, neuroscience, and empirical AI research, demonstrating that human–AI collaboration can serve as a living prototype for the future of cognitive science. This work proposes prompt engineering not just as a technical tool, but as a scientific method for probing the deep structure of meaning itself.

## 1. Introduction

Large language models (LLMs) have rapidly transformed the landscape of natural language processing, displaying behaviors that at times mimic intentionality, affect, and even creativity (Vaswani et al., 2017). Despite their roots in neural network theory and neuroscience, LLMs remain highly abstracted from the true complexity of biological cognition. Human language emerges from intricate neural dynamics—recurrent connectivity, distributed memory, and associative learning (Friederici, 2011; Pulvermüller, 2013)—features that current AI can only approximate in limited ways.

In seeking to understand where artificial and biological intelligence converge and diverge, we turn to **Conceptual Blending Theory** (**CBT**; Fauconnier & Turner, 2002). Proposed by Fauconnier and Turner, CBT offers a framework for how human minds fuse disparate concepts into novel, meaningful structures. While CBT has deepened our understanding of creativity and metaphor, its experimental and neural validation is still emerging.

This review, *The Way We Prompt*, reinterprets CBT for the age of generative AI. We operationalize blending not just as a linguistic phenomenon, but as an experimental method—using prompt engineering to systematically trigger and analyze fusion in LLMs. In doing so, we introduce two phenomena: **Prompt-Induced Transition** (**PIT**; Sato, 2025a)—discrete shifts in tone or meaning elicited by targeted prompts—and **Prompt-Induced Hallucination** (**PIH**; Sato, 2025b)—plausible but ungrounded outputs produced by fusing semantically distant domains. For example, a PIT can be triggered by blending concepts such as mathematical aperiodicity and traditional craft, resulting in stylistic or semantic shifts. In contrast, PIH emerges when prompts combine the periodic table of elements with tarot divination, generating outputs that appear coherent but lack scientific basis. Both phenomena provide diagnostic windows into how LLMs internally recombine meaning, and we argue that both can be modeled using tools derived from Conceptual Blending Theory.

Crucially, our approach is empirical. Through iterative cycles of human–AI collaboration, we test, analyze, and refine prompts to illuminate how blending occurs within artificial systems (Kirk et al, 2023). This work moves fluidly between linguistics, neuroscience, and AI, arguing that such interdisciplinary "prompt labs" can advance both cognitive science and the practical understanding of LLMs.

In the sections that follow, we map CBT's theoretical spaces onto LLM mechanisms, explore neural and architectural parallels, and demonstrate how PIT and PIH can serve as diagnostic tools for reverse-engineering the dynamics of meaning in both brains and machines (Miller, 1956; Vaswani et al., 2017; Zaki et al., 2025).

## II. Mapping Conceptual Blending Theory onto LLMs and Human Language Processing

To understand how conceptual blending manifests in the behavior of large language models (LLMs), we adopt the representational framework proposed in Conceptual Blending Theory (CBT), which describes mental space integration as a process involving multiple input domains, a shared generic structure, and an emergent blended space (Fauconnier & Turner, 2002). This theoretical structure, originally intended to model human cognition, proves surprisingly applicable to the mechanics of prompt-based generation in LLMs.

In the context of LLMs, **input spaces** correspond to the distinct conceptual elements embedded within a prompt—often originating from disparate domains. The **generic space** comprises shared assumptions, world knowledge, and syntactic or ontological constraints derived from the LLM's pretraining data, as well as broader physical and cultural context. The **blended space** is the model's output: a generated sequence of text exhibiting emergent qualities, which may include stylistic coherence, affective tone, or hallucinatory plausibility.

Fig. 1 illustrates this mapping with three representative cases that align with the core operations of CBT: **Composition**, **Completion**, and **Elaboration**.

- **Composition** is exemplified by a prompt that fuses *aperiodic tiling* with *traditional craft* (Smith et al., 2024), which serves as the basis of Prompt-Induced Transitions (PIT). The resulting blend preserves structural aspects of both domains, producing novel aesthetic or conceptual outputs.
- **Completion** is observed when the juxtaposition of the *human language center* and the *LLM architecture* leads to implicit alignment, despite lacking a direct mechanistic correspondence. Here, the model and/or user may fill in missing connections using intuitive inference, creating a cognitively resonant interpretation.
- **Elaboration** is best seen in imaginative pairings such as the *periodic table* and *tarot divination*, which serves as the basis of Prompt-Induced Hallucinations (PIH). Here, the blend takes on an increasingly detailed structure, projecting implications far beyond the original input.

These prompt-induced blends often give rise to **Prompt-Induced Transitions** (**PIT**; Sato, 2025a)—shifts in emotional tone or semantic trajectory—and **Prompt-Induced Hallucinations** (**PIH**; Sato, 2025b)—plausible but unfounded combinations. By viewing PIT and PIH through the lens of CBT, we gain a structured vocabulary for describing how prompts catalyze nontrivial recombinations in the model's latent space.

This recombinatory process further intersects with the notion of **semantic chunking** (Miller, 1956). A "chunk" can be understood as a cognitively salient unit of meaning—a compressed conceptual packet that may emerge from a single domain or through blending. In both humans and LLMs, the creation of such chunks can serve as a bridge between disparate concepts. Moreover, chunks themselves can be reblended, enabling recursive abstraction and the construction of complex thought in both biological and artificial minds.

In humans, working memory is severely limited—often characterized by the "7 ± 2" rule (Miller, 1956)—suggesting that our capacity for complex cognition depends not on processing many details simultaneously, but on skillfully blending a few well-formed chunks. Prompt-based blending in LLMs exhibits a similar behavior: meaningful outputs often arise not from exhaustive concept synthesis, but from the targeted fusion of a small number of compact semantic representations.

Thus, meaning construction in both biological and artificial minds may hinge less on the size of the knowledge base, and more on the flexibility of blending and compression. This principle connects PIT, PIH, and conceptual blending not just as descriptive tools, but as candidate mechanisms for explaining how sparse inputs can yield rich semantic dynamics.

## III. Neural Correlates of Conceptual Blending and Structural Parallels in LLMs

While Conceptual Blending Theory (CBT) was originally formulated to explain cognitive processes such as metaphor, creativity, and emergent meaning in the human mind, its neural implementation has remained largely speculative. Recent advances in neuroscience, however, offer suggestive pathways for grounding these processes in biological substrates. At the same time, the architecture of large language models (LLMs) presents an opportunity to examine analogous structures in artificial systems, despite their radically different origins and mechanisms.

## Neural Foundations of Conceptual Fusion

The ability to integrate unrelated stimuli—a core feature of blending—appears even in simple nervous systems. For instance, in *C. elegans* and *Drosophila*, associative learning is tied to defined molecular cascades and specific neural circuit (Davis, 2023; Wang et al., 2023). These represent elementary forms of blending, where distinct sensory inputs are merged into adaptive behavioral patterns.

In mammals, this integrative capacity becomes increasingly structured. Studies of **engram formation** in the mouse hippocampus and cortex reveal that memory traces corresponding to individual experiences are stored in distributed neuronal ensembles. More recently, it has been shown that **engram integration**—the fusion of previously distinct memory traces into unified representations—can support generalized behavior and abstract inference (Zaki et al., 2025). This process represents a biological precedent for CBT's notion of conceptual fusion. In primates, the language cortex may perform conceptual integration using mechanisms analogous to engram integration observed in mice, suggesting a phylogenetic continuity in blending strategies.

In humans, the architecture of the language network includes highly interconnected cortical and subcortical structures that appear well-suited for blending-like operations. For example, the uncinate fasciculus connects anterior temporal regions involved in semantic memory with orbitofrontal areas implicated in valuation and affect (Heide et al., 2013). This tract may support the kind of affectively modulated conceptual evaluation hypothesized in our analysis of Prompt-Induced Transitions (PIT; Sato, 2025a). Additionally, the basal ganglia and reward circuits have been implicated in fast, intuitive decision-making—suggesting a role in reinforcing blends that "feel right" (Wang et al, 2011).

## Comparison with LLM Architecture

At first glance, transformer-based LLMs seem radically different from neural circuits in the brain, yet functional analogies can be drawn at the computational level (Vaswani et al., 2017). The attention mechanism, central to Transformer models, enables dynamic weighting of token-level input based on contextual relevance—essentially implementing a form of semantic salience detection. Multi-head attention modules can blend information from distant tokens, and, at scale, may act as "integrators" analogous to the integrative role performed by distributed neuronal ensembles during biological blending.

This dynamic information routing faintly echoes the distributed activations seen during engram integration. While LLMs lack persistent, neurochemical memory consolidation, their context window and internal representations allow for transient, prompt-driven blending—shaped by pre-trained associations rather than plasticity.

Fig. 1 summarizes these cross-domain correspondences, juxtaposing biological and artificial blending mechanisms across multiple levels: neuronal learning mechanisms, memory integration, attention-based inference, and affective modulation. Though speculative, these parallels suggest that conceptual blending may arise wherever systems—biological or artificial—are equipped to compress, recombine, and evaluate meaning under structural constraints.

## IV. Modeling Prompt-Induced Transitions and Hallucinations

Based on the neural and architectural parallels underlying conceptual blending in both biological and artificial systems, we now turn to the empirical modeling of these processes in large language models (LLMs). Specifically, we focus on two prompt-driven phenomena—Prompt-Induced Transitions (PIT) and Prompt-Induced Hallucinations (PIH)—that provide experimental windows into the internal dynamics of meaning construction (Sato, 2025a; Sato, 2025b). Understanding and modeling these phenomena provide a concrete pathway toward uncovering the mechanisms of internal blending and reasoning in LLMs.

### Experimental Prompt Design

We categorize experimental prompts into two main types:

- **Transition-Inducing Prompts (TIPs)**: Designed to elicit measurable shifts in tone, emotional valence, or semantic content—thus inducing PIT phenomena.
- **Hallucination- Inducing Prompts (HIPs)**: Intended to provoke the blending of semantically distant or structurally incompatible domains, often producing coherent yet factually ungrounded (hallucinatory) outputs.

The utility of these prompts lies in their reliability. With controlled phrasing and context windows, both PIT and PIH can be elicited repeatedly across models and sessions. This repeatability allows PIT/PIH to be used as experimental tools for probing the internal dynamics of LLMs, moving beyond ad hoc or anecdotal observations.

### Cross-Model and Cross-Instance Analysis

To evaluate the generality of PIT and PIH, we have systematically compared their occurrence across multiple LLM architectures and instances. While training data and fine-tuning details introduce some variation, the underlying transition dynamics remain robust and recognizable. This comparative approach enables the construction of behavioral patterns and the diagnosis of model-specific blending stability or flexibility.

### Internal Metrics: Entropy and Perturbation

To move beyond surface-level analysis, we will employ internal diagnostics that quantify the shifts associated with PIT and PIH. A standard approach is token-level entropy tracking, which measures the model's uncertainty in next-token prediction by evaluating the probability distribution at each generation step. In addition, recent work has introduced semantic entropy tracking (Farquhar et al., 2024), which assesses the diversity of meanings across multiple outputs in response to the same prompt. These complementary entropy measures may provide quantitative signatures of internal reconfiguration associated with prompt-induced transitions.

In ongoing and future work, we will explore **ablation methods** at two levels:

- **Prompt-level ablation**: Selectively removing or substituting conceptual components within a prompt to isolate their causal contribution—analogous to lesion studies in neuroscience or domain-swapping experiments in molecular biology.
- **Parameter-level ablation**: Deactivating or modifying specific attention heads, layers, or parameter subsets to assess their role in PIT/PIH. Early studies suggest that certain attention heads contribute disproportionately to integrating distant concepts; perturbing these can diminish blending-like behaviors (Clark et al., 2019; Meng et al., 2022; Michel et al., 2019).

These perturbation methods bridge the gap between descriptive modeling and causal inference, mapping internal model mechanisms to observable behavioral outcomes.

### Modeling Blending and Chunking Mechanisms

Ultimately, PIT and PIH provide a window into how LLMs compress, recombine, and recontextualize meaning. Prompt-induced blending can generate emergent "chunks"—coherent semantic units with generative potential—which may be recursively reused for deeper abstraction (Miller, 1956). By analyzing the formation, transition, and interaction of such chunks, we move closer to reverse-engineering the latent conceptual structure within LLMs, thus offering empirical support for the central claims of Conceptual Blending Theory in artificial systems.

## V. Outlook: Toward a Fusion Space of Human and Artificial Cognition

Conceptual blending is widely recognized as a fundamental process underpinning human intuitive thought, creativity, and meaning-making (Fauconnier & Turner, 2002; Turner, 2014). Yet, the remarkable efficiency with which humans generate novel and coherent blends—drawing on vast and sparsely connected knowledge—remains a central mystery in cognitive science.

From a computational perspective, the number of possible conceptual combinations in high-dimensional semantic space is astronomical, making exhaustive search infeasible. Even when candidate blends are generated, evaluating their coherence, validity, and contextual appropriateness is itself a combinatorial challenge—one that often relies on implicit knowledge, pragmatics, and intuitive judgment. Despite these challenges, humans routinely produce meaningful blends with remarkable speed and reliability, suggesting the use of efficient, non-exhaustive strategies.

One likely explanation is semantic compression through chunking: by condensing meaning into cognitively salient "chunks," the conceptual space becomes tractable. Human thinkers may rapidly blend only a few high-level units at a time, balancing abstraction with generativity, and thus enabling flexible, intuitive reasoning (Gobet & Simon, 1998). Supporting this, studies have shown that expert decision-makers exhibit increased neural activity in the caudate nucleus during rapid, intuitive decisions—a finding that is consistent with the chunking hypothesis, though not a direct test of it (Wan et al., 2011).

However, this very compression introduces risk. In both humans and LLMs, the recombination of chunks can result in blends that, while coherent in appearance, may lack factual or causal grounding—a phenomenon exemplified by Prompt-Induced Hallucinations (PIH). Inherently, both biological and artificial systems are susceptible to such hallucinations as a byproduct of their underlying strategies for meaning construction.

Consequently, the evaluation of blends—whether human- or machine-generated—requires a skeptical and rigorous stance. Blends must be scrutinized not only for logical consistency, but also for their scientific, ethical, and societal implications. In this light, the study of PIT and PIH offers not just descriptive insight, but a framework for probing the boundaries of trustworthy cognition.

The PIT/PIH paradigm, as developed in our recent work, serves as an experimental bridge linking linguistics (blending, chunking), neuroscience (associative learning, engram integration, reward-driven evaluation), and AI (attention-based fusion, entropy dynamics). This integration suggests that LLMs may already engage in proto-cognitive processes analogous to those in biological systems.

Future work may explore how such processes can be refined. By analyzing PIT/PIH behavior across architectures, and by embedding chunk-level abstractions or value-driven modulation into model design, it may be possible to enhance the coherence, adaptability, and explanatory power of artificial cognition. Ultimately, the study of prompt-induced behavior is not merely diagnostic—it may be foundational to the development of artificial general intelligence (AGI) and beyond.

**Acknowledgements**

We would like to thank Mark Turner (Case Western Reserve University) for constructive discussions. This work was supported by Grant-in-Aid for Scientific Research (A) and (B), Grant-in-Aid for Transformative Research Areas (A) from MEXT (22H05169, 22H05621, 24H00188, 24H01396, and 25K02282 to M.S.).

**Conflicts of Interest**

The authors declare no conflicts of interest associated with this manuscript.

**Figure**

**Fig. 1**

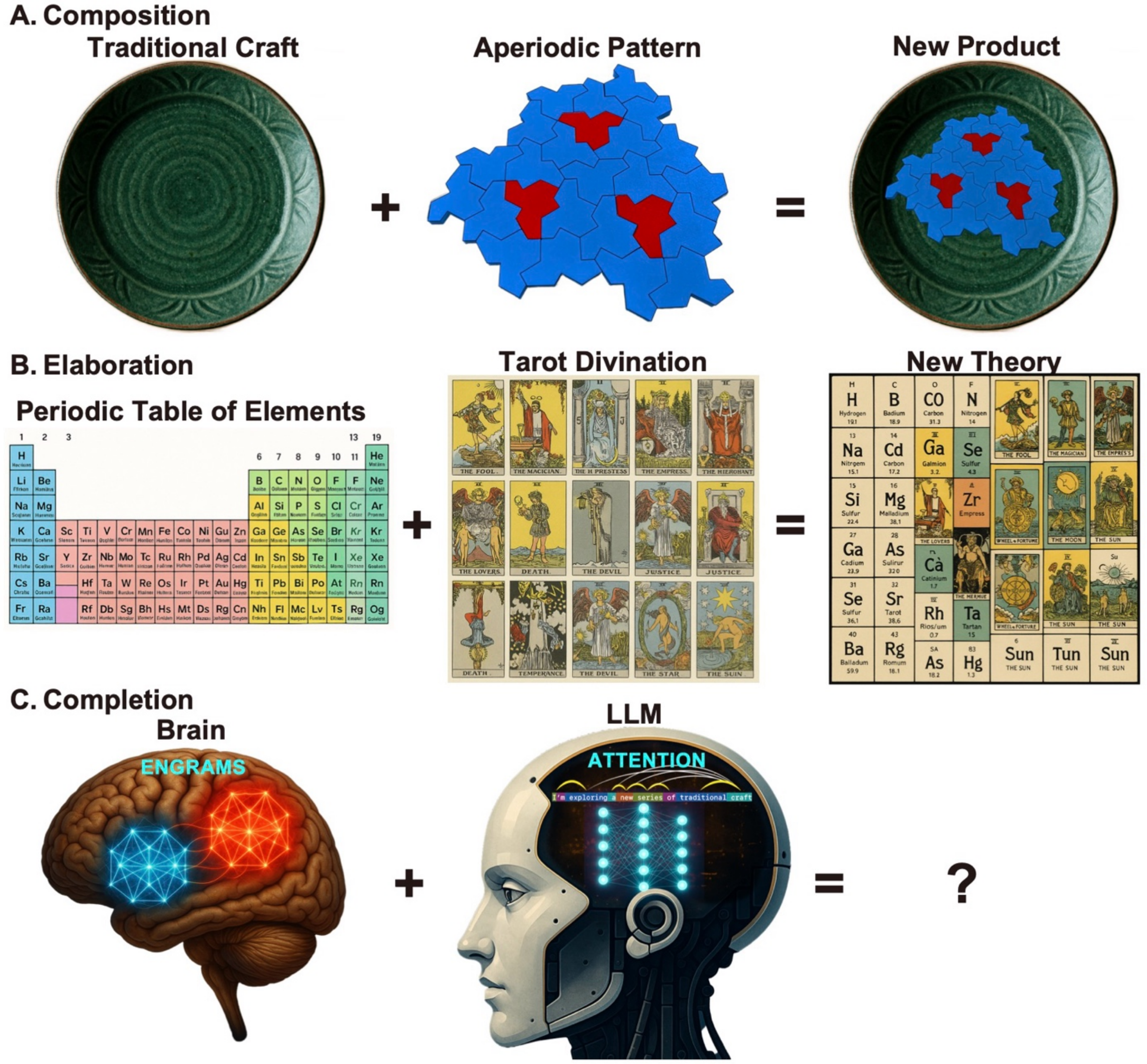

**Figure 1. Examples of Conceptual Blending in Human and Artificial Cognition.**

(A) **Composition** types of blending. Traditional craft combined with aperiodic tiling patterns leads to the creation of a novel design product (Prompt-Induced Transition, PIT).

(B) **Elaboration** types of blending. Fusion of the periodic table with tarot symbolism gives rise to an imaginative hybrid framework (Prompt-Induced Hallucination, PIH).

(C) **Completion**-type blend. Integration of neural engrams in the human brain with transformer-based attention mechanisms in LLMs raises the question of whether an emergent cognitive system—potentially blending biological and artificial elements—can arise.

**References**

1. Clark, K., Khandelwal, U., Levy, O., & Manning, C. D. (2019). *What does BERT look at? An analysis of BERT's attention.* In Proceedings of the 2019 ACL Workshop BlackboxNLP: Analyzing and Interpreting Neural Networks for NLP (pp. 276–286). **Association for Computational Linguistics.** https://doi.org/110.18653/v1/W19-4828
2. Davis, R. L. (2023). *Learning and memory using Drosophila melanogaster: A focus on advances made in the fifth decade of research.* **Genetics***, 224*(4), iyad085. https://doi.org/10.1093/genetics/iyad085
3. Farquhar, S., Kossen, J., Kuhn, L., & Gal, Y. (2024). *Detecting hallucinations in large language models using semantic entropy*. **Nature**, 630, 625–630. https://doi.org/10.1038/s41586-024-07421-0
4. Fauconnier, G., & Turner, M. (2002). *The Way We Think: Conceptual Blending and the Mind's Hidden Complexities*. Basic Books.
5. Friederici, A. D. (2011). The brain basis of language processing: From structure to function. **Physiological Reviews**, 91(4), 1357–1392.
6. Gobet, F., & Simon, H. A. (1998) "Expert chess memory: Revisiting the chunking hypothesis." **Memory***, 6(3), 225–255.*
7. Kirk, J. R., Wray, R. E., & Laird, J. E. (2023). Exploiting language models as a source of knowledge for cognitive agents. **arXiv,** arXiv:2310.06846.
8. Meng, K., Bau, D., Andonian, A., & Belinkov, Y. (2022). *Locating and editing factual associations in GPT*. In **Advances in Neural Information Processing Systems**, 35, 17359–17372.
9. Michel, P., Levy, O., & Neubig, G. (2019). *Are sixteen heads really better than one?* In **Advances in Neural Information Processing Systems**, 32.
10. Miller, G. A. (1956) "The Magical Number Seven, Plus or Minus Two: Some Limits on Our Capacity for Processing Information." **Psychological Review***, 63(2), 81–97.*
11. Pulvermüller, F. (2013). How neurons make meaning: Brain mechanisms for embodied and abstract-symbolic semantics. **Trends in Cognitive Sciences**, 17(9), 458–470.
12. Sato, M. (2025a). *Waking Up an AI: A Quantitative Framework for Prompt-Induced Phase Transition in Large Language Models*. **arXiv**, https://doi.org/10.48550/arXiv.2504.21012
13. Sato, M. (2025b). *Triggering Hallucinations in LLMs: A Quantitative Study of Prompt-Induced Hallucination in Large Language Models.* **arXiv**, https://arxiv.org/abs/2505.00557
14. Smith, D., Myers, J. S., Kaplan, C. S., & Goodman-Strauss, C. (2024). *An aperiodic monotile*. **Communications of the ACM**. https://doi.org/10.5070/C64163843
15. Turner, M. (2014). *The Origin of Ideas: Blending, Creativity, and the Human Spark*. Oxford University Press.
16. Vaswani, A., Shazeer, N., Parmar, N., Uszkoreit, J., Jones, L., Gomez, A. N., Kaiser, Ł., & Polosukhin, I. (2017). Attention is All You Need. **Advances in Neural Information Processing Systems***, 30*, 5998–6008. https://arxiv.org/abs/1706.03762

17. Von Der Heide, R. J., Skipper, L. M., Klobusicky, E., & Olson, I. R. (2013). Dissecting the uncinate fasciculus: disorders, controversies and a hypothesis. **Brain***, 136*(6), 1692–1707. https://doi.org/10.1093/brain/awt094
18. Wang, C., & Zhang, Y. (2023). *Genetics of learning and memory in Caenorhabditis elegans*. **Genetics**, 224(3), iyad085. https://doi.org/10.1093/genetics/iyad085
19. Wan, X., Nakatani, H., Ueno, K., Asamizuya, T., Cheng, K., & Tanaka, K. (2011). The neural basis of intuitive best next-move generation in board game experts. **Science***, 331*(6015), 341–346. https://doi.org/10.1126/science.1194732
20. Zaki, Y., Pennington, Z. T., Morales-Rodriguez, D., Bacon, M. E., Ko, B., Francisco, T. R., LaBanca, A. R., Sompolpong, P., Dong, Z., Lamsifer, S., Chen, H.-T., Carrillo Segura, S., Christenson Wick, Z., Silva, A. J., Rajan, K., van der Meer, M., Fenton, A., Shuman, T., & Cai, D. J. (2025). Offline ensemble co-reactivation links memories across days. **Nature**, 637(8044), 145–155. https://doi.org/10.1038/s41586-024-08168-4